\documentclass{article}

\usepackage{microtype}
\usepackage{graphicx}
\usepackage{subfigure}
\usepackage{booktabs} 
\usepackage{multirow}

\usepackage{hyperref}

\usepackage[accepted]{icml2024}

\usepackage{amsmath}
\usepackage{amssymb}
\usepackage{mathtools}
\usepackage{amsthm}

\usepackage[capitalize,noabbrev]{cleveref}

\theoremstyle{plain}

\theoremstyle{definition}

\theoremstyle{remark}

\usepackage[textsize=tiny]{todonotes}

\icmltitlerunning{Unveiling the Potential of AI for Nanomaterial Morphology Prediction}

\begin{document}

\twocolumn[
\icmltitle{Unveiling the Potential of AI for \\
           Nanomaterial Morphology Prediction}




\begin{icmlauthorlist}
\icmlauthor{Ivan Dubrovsky}{ITMO}
\icmlauthor{Andrei Dmitrenko}{ITMO,DONE}
\icmlauthor{Aleksei Dmitrenko}{ITMO}
\icmlauthor{Nikita Serov}{ITMO}
\icmlauthor{Vladimir Vinogradov}{ITMO}
\end{icmlauthorlist}

\icmlaffiliation{ITMO}{Center for AI in Chemistry, ChemBio Cluster, ITMO University, St. Petersburg, Russia}
\icmlaffiliation{DONE}{D ONE AG, Zurich, Greater Zurich Area, Switzerland}

\icmlcorrespondingauthor{Vladimir Vinogradov}{vinogradov@scamt-itmo.ru}
\icmlcorrespondingauthor{Andrei Dmitrenko}{dmitrenko@scamt-itmo.ru}

\icmlkeywords{Machine Learning, ICML, Nanomaterials, Morphology, LLMs, Image Generation}

\vskip 0.3in
]



\printAffiliationsAndNotice{}  

\begin{abstract}
Creation of nanomaterials with specific morphology remains a complex experimental process, even though there is a growing demand for these materials in various industry sectors. This study explores the potential of AI to predict the morphology of nanoparticles within the data availability constraints. For that, we first generated a new multi-modal dataset that is double the size of analogous studies. Then, we systematically evaluated performance of classical machine learning and large language models in prediction of nanomaterial shapes and sizes. Finally, we prototyped a text-to-image system, discussed the obtained empirical results, as well as the limitations and promises of existing approaches.

\end{abstract}

\section{Introduction}
\label{Introduction}
Nowadays, nanomaterials are spread across many fields of science and industry \cite{zebarjadi_low-temperature_2011, liu_potentials_2015, kairdolf_bioconjugated_2017, shifrina_role_2020, gao_encapsulated_2021, takechi-haraya_current_2022}. In each of those fields, for a nanomaterial to be fit for purpose, its size, shape, and other morphological parameters must be precisely controlled, as they directly influence toxicity, catalytic activity and other properties of nanomaterials crucial for applications. Altering these parameters also allows to improve efficiency of drug delivery systems \cite{sen_gupta_role_2016}, catalysts \cite{shifrina_role_2020}, energy storage systems \cite{pomerantseva_energy_2019}, etc.

Typically, creating a nanomaterial with a specific set of properties requires a significant number of experiments ranging from a few repetitive syntheses to a dozen of substantially different synthesis procedures \cite{vaidyanathan_characterization_2008,sun_co-precipitation_2021}. Each synthesis is followed by a specific method of analysis to confirm the experimental outcome. One of the most prominent methods for analyzing nanomaterials is the scanning electron microscopy (SEM) \cite{smith_scanning_1955}. With SEM, it is possible to obtain information about the size and shape of nanoparticles (NPs), as well as the structure of the surface, surface flaws and contaminants. Currently, the SEM method is deemed irreplaceable despite being costly and time-consuming \cite{singh_experimental_2016}. On average, one analysis with SEM can cost up to a few hundred US dollars, leading to vast amounts of resources required to run any large-scale study. While conducting such experiments scientists are most often guided by their experience and intuition acquired in past experiments. This is because the problem of determining the morphology of nanomaterials based on the synthesis parameters currently has no theoretical or computational solution, in general. At the same time, syntheses of nanomaterials include too many different interdependent parameters for a person to be able to account for. Therefore, there is a high demand \cite{91de84b9ab31471ba1090a656b0d8527} for predictive models capable of characterizing the properties of nanomaterials bypassing the need of costly experimental work.

Artificial intelligence (AI) offers the most promising set of tools to meet this demand. In fact, classical machine learning (ML) models including artificial neural networks have already been successfully applied to many tasks related to nanomaterial science \cite{serov_artificial_2022,chen_ai_2023,banaye_yazdipour_predicting_2023}. With recent astonishing advances in deep learning \cite{jumper_highly_2021,rombach_high-resolution_2021,ramesh_hierarchical_2022,openai_gpt-4_2023,touvron_llama_2023, jiang2023mistral, merchant2023scaling}, the potential of AI in the design of nanomaterials seems truly immense. However, one has to possess large volumes of carefully curated data to fully exploit the power of AI. As discussed earlier, accumulating the data appropriate for the prediction of nanomaterial morphology has been a major challenge for decades. Within realistic data constraints, the boundaries of AI in the design of nanomaterials are underexplored.

One of the goals of this work was to showcase possible applications of the most recent advances in machine learning to the design of nanomaterials, and bridge the gap between the experimentalists and the machine learning experts. A huge number of scientific groups are engaged in optimization of various parameters of nanomaterials, in particular morphology \cite{SHANDILYA2022128218, doi:10.1126/sciadv.abo2626, nano12122087, https://doi.org/10.1002/ppsc.202300118}, as it allows new industrial applications or improves the properties of existing materials. The importance of this direction of research is also highlighted by the recent efforts of the leading AI companies, such as DeepMind \cite{merchant2023scaling}.

In this study, we aim to unveil the capabilities and limitations of AI in predicting morphology of nanomaterials. For that, we first conduct 215 experimental syntheses of calcium carbonate-based nanomaterials of different shapes and sizes. We carefully document the synthesis procedures with the parameters of experimental conditions, take SEM-images of the resulting nanoparticles, segment and manually annotate them with expert knowledge. We investigate the statistical associations in this multimodal dataset and identify features informative of nanoparticle morphology. We further use these findings to train classical ML models to predict sizes and shapes of nanoparticles and achieve 0.77 and 0.80 average accuracy, respectively. For the first time in the field of nanomaterial synthesis, we explore the potential of LLMs for prediction tasks. Using few-shot methods, we utilize state-of-the-art models, such as GPT-4, to predict the shapes of nanomaterials and achieve an impressive 0.81 average accuracy. Finally, we augment the available data to prototype a text-to-image system aimed at generating an image of a nanoparticle based on the description of its synthesis procedure. In conclusion, we review the obtained empirical results and discuss the future of AI in the field of nanomaterial design. 


\section{Related work}
Over the past 10 years, there have been several works predicting morphological properties of nanoparticles. However, the majority of them focused on size prediction considering a single experimental system, where the resulting particles conform to the same shape and their sizes can be easily standardized. Some particular examples include size prediction for silver nanoparticles \cite{chen_prediction_2016, shafaei_predictive_2020}, carbon nanotubes \cite{iakovlev_artificial_2019}, agar nanospheres \cite{zaki_preparation_2015}, chitosan nanoparticles \cite{baharifar_size_2017}, polymeric nanoparticles \cite{shahsavari_application_2013,soliman_determination_2014,youshia_artificial_2017}, TiO\textsubscript{2} nanoparticles \cite{pellegrino_machine_2020} and different methacrylates \cite{kimmig_prediction_2021}. In our work, there is no attachment to nanoparticles of a certain shape. Instead, we generate a dataset containing multiple different shapes, which greatly expands the generalizability of our approach and enables future transfer learning applications. In addition, unlike many previous studies, we provide the data for benchmarking and the code for reproducibility. 

A few published works specialize in predicting the shapes of nanoparticles \cite{timoshenko_supervised_2017,chen_machine-learning-guided_2020,yao_seeking_2022}, but they too have certain shortcomings. For example, Timoshenko et al. created a model that takes experimental X-ray absorption near-edge structure (XANES) spectroscopy data as input to predict the 3D structure of metallic nanoparticles \cite{timoshenko_supervised_2017}. Although circumventing the need for SEM analysis, this approach still requires actual synthesis and experimental evaluation of other properties to predict the shape of the nanomaterial. This narrows down the list of possible applications significantly. In contrast, our work explores data-driven approaches that only use features of the past syntheses to predict morphology of potentially new nanomaterials.

More advanced deep learning algorithms have also found applications in the creation of new nanomaterials \cite{roccapriore_predictability_2021, xu_machine_2023}. In the paper by Kim, Han, and Han, a model based on convolutional neural networks was proposed capable of determining the morphology of nanomaterials based on the SEM images \cite{kim_machine_2020}. Such efforts help to better understand morphological properties of nanomaterials and simplify data labeling for the future predictive approaches. However, they do not avoid tedious experimental work preparing the datasets of SEM images, by design. Ultimately, our work stands out by predicting SEM images of nanoparticles of different morphologies based on the properties of the corresponding syntheses, which is an inverse problem formulation.

Recent advances in natural language processing \cite{openai2023gpt4, jiang2023mistral, touvron_llama_2023} have also been reflected in some areas of chemistry. Recently, there have been studies that describe the use of LLMs, in particular using the few-shot method, to predict the characteristics of various chemical objects \cite{zheng2023gpt} and even to generate new chemical structures \cite{jablonka2022is}. However, the potential application of LLMs to predict the morphology of nanomaterials has not yet been investigated.

Various multimodal systems have been proposed recently in application to nanomaterial science \cite{kononova_text-mined_2019, lee_statistical_2020, hiszpanski_nanomaterial_2020}. Since the emergence of Stable Diffusion \cite{rombach_high-resolution_2021} and DALL-E \cite{ramesh_hierarchical_2022}, image generation models have attracted particularly much public attention. A recent work in nanofabrication presented an image-to-image system capable of predicting the postfabrication appearance of structures manufactured by focused ion beam milling \cite{buchnev_deep-learning-assisted_2022}. Although a very specialized application, it demonstrates how the field of nanotechnology already benefits from generative AI. In this work, we prototyped a text-to-image solution predicting morphologies of the previously unseen nanomaterials.

\section{Dataset preparation}
To obtain the most reliable and standardized dataset, we performed 215 syntheses of calcium carbonate-based nanomaterials. As mentioned above, usually up to several dozen experiments are conducted to perform optimization of nanomaterial properties. When using machine learning, more samples are usually needed to build predictive models, but due to the resource-intensive and time-consuming nature of synthesizing and analyzing nanomaterials, most of the morphology prediction works described above are limited to about a hundred syntheses. In this work, we generated a dataset that is double that size. 

We considered a single chemical system of calcium carbonate, because of its rich variety of nanoparticle shapes and sizes. By making this study design choice, we were hoping to achieve better generalization of our work to other nanoparticles, since most of the known shapes are already represented in our dataset.

For each synthesis, we documented all variable parameters, such as names of reagents, solvents, etc., their concentrations, temperature and reaction time, as well as other synthesis parameters. Additionally, for each synthesis, one most representative SEM-image was taken, which clearly shows nanoparticles with distinguishable sizes and shapes.

We thoroughly analyzed shapes and sizes of the resulting nanoparticles and identified five different shapes: cubic, spherical, stick-shaped, flat, and amorphous. For each shape, except flat and amorphous, we distinguished small-, medium- and large-sized nanoparticles applying an empirical threshold. In the case of amorphous and flat particles, the number of samples was too small to consider differentiation. Altogether, we used 5 different shape categories and 9 different categories combining shapes and sizes to label the dataset.

To train the variational autoencoder in the text-to-image setup described later, each image from the original dataset with 215 syntheses was segmented to extract multiple images of individual nanoparticles using ImageJ \cite{rueden_imagej2_2017}. The resulting dataset was further augmented to increase the size of the dataset and decrease the probability of overfitting. For that, we generated new images by applying random rotations, different blurring and brightness settings. In total, the training dataset contained 46,800 images of individual nanoparticles.

\section{Feature selection}
Each synthesis in our dataset was described by 10 continuous and 3 categorical variables that might be influencing the shapes of nanomaterials in different ways. This section describes statistical evaluation of those features to determine whether they are indeed informative of the geometry of nanomaterials, which served as a basis for downstream AI applications.

\subsection{Analysis of continuous variables}

Let ($X_{1}^{1}$, $X_{2}^{1}$, …, $X_{n}^{1}$) denote real values of a parameter of a synthesis which produces cubic nanoparticles. Let ($X_{1}^{2}$, $X_{2}^{2}$, …, $X_{m}^{2}$) denote the real values of the same parameter of any synthesis which always results in nanoparticles of different shapes. We wondered whether the two samples came from the same population or not. If so, each value of the first sample would have had an equal chance of being larger than each value of the second sample. Therefore, the null hypothesis can be formulated as follows: 

\[H_{0}:p(X_{i}^{1} > X_{j}^{2})= \frac{1}{2}\]

In fact, this formulation represents the Mann-Whitney U test \cite{nachar_mann-whitney_2008}. We applied it for each of the real-valued parameters of synthesis and each type of the nanomaterial shapes. We found that formation of stick-shaped nanoparticles was dependent on the reaction temperature, synthesis time, and polymer mass and/or concentration. Cubic shapes of nanoparticles were also associated with certain temperatures and polymer concentrations, as well as the molar mass of the polymer. We used the Kruskal-Wallis H test \cite{kruskal_use_1952}, which is analogous to Mann-Whitney U test but applicable to three and more sample groups,  Kolmogorov-Smirnov test \cite{smirnov_estimate_1939} and ANOVA \cite{marsal_introduction_1987} to corroborate these findings. Herewith, we used the significance level = 0.05 and the Bonferroni correction method to account for multiple hypothesis testing.

\subsection{Analysis of categorical variables}

To establish relationships between categorical parameters of synthesis procedures and the corresponding shapes of nanomaterials, we composed contingency tables as shown in \autoref{table_contingency}.

\begin{table}[h]
		\caption{Example contingency table for testing categorical variables of synthesis procedures.}
		\label{table1}
    \vskip 0.15in
    \begin{center}
    \begin{small}
	\centering
		\begin{tabular}{@{}l c c@{}}
  \toprule
			& Compound & Compound \\ 
			& in synthesis & not in synthesis \\
			\midrule
			NPs of a given shape & \textit{a} & \textit{b}  \\ 
			NPs of other shapes  & \textit{c} & \textit{d}  \\ 
			\bottomrule
		\end{tabular}
    \end{small}
    \end{center}
    \vskip -0.1in
    \label{table_contingency}
\end{table}

According to Fisher, $a \sim Hypergeometric(N, K, n)$, where $N = a + b + c + d$ is the population size, $K = a + b$ is the number of successes and $n = a + c$ is the number of draws \cite{fisher_interpretation_1922}. Therefore, the probability of this outcome is given by: 

\[
	\begin{aligned}
		p=\frac{\begin{pmatrix}
		a+b\\ 
		a
	\end{pmatrix}
	\begin{pmatrix}
		c+d\\ 
		c
\end{pmatrix}}{\begin{pmatrix}
		n\\ 
		a+c
\end{pmatrix}} = 
\frac{\begin{pmatrix}
		a+b\\ 
		b
	\end{pmatrix}
	\begin{pmatrix}
		c+d\\ 
		d
\end{pmatrix}}{\begin{pmatrix}
		n\\ 
		b+d
\end{pmatrix}} = \\
= \frac{(a+b)!(c+d)!(a+c)!(b+d)!}{a!b!c!d!n!}
  \end{aligned}
\]
We computed these probabilities for each combination of nanoparticle shape and polymer/surfactant/solvent involved in the synthesis. Using the same significance level and the correction for multiple hypothesis testing as before, we observed several strong associations: stick-shaped nanoparticles with polyethylene glycol (PEG) and polyethylenimine (PEI) polymers; flat nanoparticles with presence of PEDOT:PSS and polyvinylpyrrolidone (PVP); cubic nanoparticles with presence of polyacrylic acid (PAA) and PEDOT:PSS. We also found strong dependencies of nanoparticles’ shapes on the following surfactants: Myristyltrimethylammonium bromide and Sodium dodecylsulfate. In the case of amorphous nanoparticles, the presence of Propylene glycol and tert-Butanol solvents was also found significant. Finally, we applied the Chi-squared test \cite{magnello_karl_2005} to confirm the aforementioned findings. For more information on how the statistical tests and the most significant associations between particular synthesis parameters and nanomaterial shapes, see Appendix~\ref{appendix_statistics}.

Notably, many of the parameters of syntheses had no effect on the shapes of nanomaterials, e.g., stirring speed, concentrations of Ca and CO\textsubscript{3} ions, presence of Hexadecyltrimethylammonium bromide and Triton X-100 surfactants, and 1-Hexanol and Methyl alcohol solvents. For the downstream machine learning applications, we excluded those features from the data.

\section{Shape and size prediction}

Statistical tests proved certain associations between the parameters of  syntheses and the morphologies of the resulting nanomaterials. Therefore, we attempted to exploit them in predicting shapes and sizes of nanomaterials using classical machine learning algorithms.
 
In some cases, several nanoparticles of different shapes and sizes were present on the same image, so initial 215 syntheses produced 314 training examples of nanoparticles of different types. Following the logic of the statistical evaluation, we formulated a set of binary classification tasks, one for each type of shape or a combination of shape and size. In this formulation, we first trained a separate model to distinguish nanoparticles of each particular shape. Then, we ran multiple predictions for each sample during the inference to establish what shapes of nanoparticles were present on the corresponding image. The same logic applied to combinations of shapes and sizes. Notably, some syntheses consistently result in nanoparticles of several different shapes. Our approach allows dealing with such ambiguities without the need to determine the prevailing nanomaterial shape or size.

\subsection{Classical machine learning}

\subsubsection{Tree-based ensemble models}

We trained the tree-based models, namely Random Forest (RF) and Gradient Boosted Trees (XGB), to predict 9 categories representing combinations of shapes and sizes and 5 categories representing shapes only. Therein, we followed all the good practices in data preprocessing and model selection. A thorough description of the process of development, optimization and evaluation of classical machine learning models is presented in the Appendix~\ref{appendix_ensemble_models}.

\subsubsection{Results}
The accuracy and the F1 scores of the best models evaluated on the test dataset are presented in \autoref{table_rf_shapes} and \autoref{table_rf_sizes}. 
Each experiment was performed 5 times at different random states, and the mean value and standard deviation were calculated.

\begin{table}[h]
	\caption{Prediction of shapes. Top average accuracy and F1 scores achieved by the Random Forest classifiers on the test set.}
	\label{table2}
    \begin{center}
    \begin{small}
	\begin{tabular}{@{}lccccc@{}}
 \toprule
		Shape & \# samples & Accuracy & F1 score \\ 
 \midrule
Cube      & 140 & 0.76 ± 0.02 & 0.73 ± 0.03 \\
Stick     & 84  & 0.78 ± 0.01 & 0.77 ± 0.01 \\
Sphere    & 40  & 0.82 ± 0.06 & 0.67 ± 0.08 \\
Flat      & 16  & 0.82 ± 0.11 & 0.52 ± 0.09 \\
Amorphous & 34  & 0.80 ± 0.02 & 0.62 ± 0.04 \\        \midrule
Average & &   0.80 ± 0.04   &     0.66 ± 0.05            \\
\bottomrule
	\end{tabular}
    \end{small}
    \end{center}
    \label{table_rf_shapes}
\end{table}

Based on our results, the shapes of nanoparticles can be predicted reasonably well (\autoref{table_rf_shapes}). For every nanomaterial shape, RF performed better than XGB, so only RF metrics are displayed. The average accuracy and F1 score were 0.80 and 0.66, respectively. Unsurprisingly, the samples of the least represented categories (namely, flat and amorphous shapes) produced lower F1 scores, which decreased the overall metrics.

\begin{table*}[tb]
	\caption{Prediction of shapes and sizes with tree-based ensemble models. Average accuracy, and F1 scores for Random Forest (RF) and Gradient Boosting (XGB) classifiers on the test set.}
    \label{table3}
    \vskip 0.15in
    \begin{center}
    \begin{small}
	\begin{tabular}{@{}lccccc@{}}
 \toprule
\multirow{2}{*}{Shape \& size} & \multirow{2}{*}{\# samples} & \multicolumn{2}{c}{Accuracy} & \multicolumn{2}{c}{F1 score} \\ \cmidrule{3-6} 
		     &    & XGB                    & RF          & XGB                   & RF          \\ \midrule
Cube\_small   & 25 & \textbf{0.85 ± 0.01} & 0.82 ± 0.04 & \textbf{0.58 ± 0.07}  & 0.57 ± 0.08 \\
Cube\_medium  & 49 & 0.64 ± 0.05 & \textbf{0.64 ± 0.03} & 0.48 ± 0.05            & \textbf{0.52 ± 0.06} \\
Cube\_large   & 66 & 0.67 ± 0.04 & \textbf{0.70 ± 0.03} & 0.61 ± 0.02            & \textbf{0.64 ± 0.03} \\ \midrule
Stick\_small  & 30 & \textbf{0.83 ± 0.03} & 0.82 ± 0.03 & 0.52 ± 0.04            & \textbf{0.54 ± 0.04} \\
Stick\_medium & 28 & \textbf{0.83 ± 0.06} & 0.81 ± 0.07 & \textbf{0.61 ± 0.07}   & 0.59 ± 0.09 \\
Stick\_large  & 26 & \textbf{0.79 ± 0.04} & 0.79 ± 0.04 & \textbf{0.64 ± 0.05}   & 0.63 ± 0.05 \\ \midrule
Sphere\_small  & 11 & \textbf{0.70 ± 0.36} & 0.68 ± 0.34 & 0.37 ± 0.19           & \textbf{0.37 ± 0.18} \\
Sphere\_medium & 19 & \textbf{0.86 ± 0.04} & 0.84 ± 0.07 & \textbf{0.55 ± 0.06}  & 0.55 ± 0.10 \\
Sphere\_large & 10 & \textbf{0.72 ± 0.27} & 0.61 ± 0.28 & \textbf{0.44 ± 0.16}   & 0.40 ± 0.18 \\ \midrule
 Average &  &   \textbf{0.77 ± 0.10}        &   0.75 ± 0.10   &      \textbf{0.53 ± 0.08}     &     0.53 ± 0.09            \\
\bottomrule
	\end{tabular}
\end{small}
\end{center}
\vskip -0.1in
\label{table_rf_sizes}
\end{table*}

Extending the number of categories to include the sizes of nanoparticles as well resulted in superior performance of XGB in most cases (\autoref{table_rf_sizes}). The overall average accuracy for the task was 0.77, and the average F1 score -- 0.53 . This drop in performance was expected, as the number of samples per category became smaller, increasing the risk of overfitting. Underrepresentation becomes even more apparent as well for some classes. Apart from evaluating the models on the test set, which had never been used during training, we also explored feature importances as an additional validation step. In most cases, we observed that the top 5 most important parameters were well in agreement with the statistical tests described in the previous section and presented in \autoref{table_statistics} of the Appendix~\ref{appendix_statistics}. An example of feature importance analysis for the Random Forest model predicting whether a nanoparticle belongs to a stick shape is shown on \autoref{figure_shap} of the Appendix~\ref{appendix_ensemble_models}.

Thus, we demonstrated the possibility of predicting shapes and sizes of NPs with machine learning models, confirmed by average test accuracy of 0.80 and by feature importance analysis coherent with the statistical evaluation. The trained models can already be used to predict morphological properties of new nanomaterials based on their synthesis procedures. However, with recent advances in large language models, we wondered whether similar prediction performance can be achieved with state-of-the-art LLMs in a few-shot scenario. That would allow material scientists to use natural language for prediction tasks, bypassing the need to develop and optimize complex machine learning pipelines. In the following section, we describe applications of LLMs to nanomaterial shape and size prediction.

\subsection{Large language models}

\subsubsection{Texts of synthesis procedures}
\label{subsection_synthesis_texts}

A dataset of texts describing synthesis procedures was prepared for morphology prediction with LLMs and training the text-to-image model. For that, we created a dozen of semantic templates with gaps for particular values of synthesis parameters. We leveraged the publicly available GPT-3.5 \cite{liu_summary_2023} to generate such templates based on a few examples taken from the scientific articles. Thanks to GPT's strong ability to paraphrase while maintaining the writing style, we managed to collect texts of synthesis procedures sufficiently different from each other in semantics but identical in contents (i.e., the sequence of actions and the list of relevant parameters). The paraphrasing was applied to make sure our approach is capable of handling real-world data and less prone to overfitting. Each template was validated by the experimental team. Peer review included both checking that the experiment description was accurate to maintain reproducibility and that the constructed templates were close to those commonly found in the scientific publications. We also did a practical evaluation of the generated templates to ensure that they contain all the necessary information about the syntheses. For that, we used BERT to extract features of the filled templates and predict the original synthesis parameters with a perceptron. In most cases, we achieved retrieval accuracy close to 1. We provide two examples of the generated templates in the Appendix~\ref{appendix_synthesis_texts}.

\subsubsection{Few-shot classification}
It is now known that LLMs can achieve quite high performance in domain-specific regression and classification tasks, often on par with the other widely accepted methods \cite{jablonka2022is}. In this study, we investigated applications of LLMs to nanomaterial morphology prediction. 

For this purpose, we used a few-shot method, in which we show the model only a few randomly selected samples from our training set and then prompt it to make a prediction for a test sample. In all experiments, we used a special prompt describing the task that the LLM was given. It starts as follows:

\textit{You are an expert in the synthesis of nanomaterials. You analyze the conditions for obtaining a nanomaterial and predict what particle shapes will be present in the synthesized material. There are five particle shapes: 'Cube', 'Stick', 'Sphere', 'Flat' and 'Amorphous'. A nanomaterial can contain particles of different shapes. If you cannot say exactly what it is, list the forms that have the highest probability in those conditions.}

We then appended several random examples from the training set with the corresponding true labels and a single example from the test subset to the prompt. While doing so, we varied the number of random examples $N$, the sampling method and the data format. We used from $N=2$ to $N=10$ training examples in the prompt. We experimented with two sampling strategies: \textit{i)} at least one training example belongs to the same target class as the test sample, \textit{ii)} all training examples belong to the same class as the test sample. The choice of the sampling strategies was based on practical considerations around real experiments. More specifically, strategy \textit{i)} is targeted at characterization of previously unknown shapes, while strategy \textit{ii)} follows the logic of a confirmation experiment, when a researcher only needs to confirm the presence of a particular nanomaterial shape in the synthesis. Finally, we used either of the two formats: textual (described in subsection ~\ref{subsection_synthesis_texts}) or tabular. In the tabular format, features of the training examples were concatenated to a string along with their values separated by colon, e.g., "Ca ion, mM: 44; CO3 ion, mM: 159...". Finally, the LLM was instructed to produce the list of nanoparticle shapes corresponding to the test synthesis as an answer. A more detailed description of prompts is presented in the Appendix~\ref{appendix_synthesis_texts}.

Using the above prompt structure, we applied 6 state-of-the-art LLMs, including GPT-4-turbo (\texttt{gpt-4-0125-preview}), GPT-4 (\texttt{gpt-4-0613}) and GPT-3.5-turbo (\texttt{gpt-3.5-turbo-1106}) from OpenAI \cite{openai_gpt-4_2023}, as well as the latest versions of Mistral Medium, Small and Tiny from Mistral AI \cite{jiang2023mistral}, to the same classification tasks described earlier. To systematically evaluate performance, we repeated each computational experiment 5 times and calculated mean and standard deviation for the standard classification metrics. 

\subsubsection{Results}

\autoref{table_llms_comparison} shows top performance of LLMs predicting shapes of nanomaterials. Strikingly, GPT-4 achieved an even higher average accuracy than tree-based ensemble models. Among the other LLMs, it also demonstrated the smallest standard deviation, which speaks for better consistency. Interestingly, the second best model was Mistral-small. Given that its inference time and pricing are much lower than GPT-4, this model could be a pragmatic choice for practitioners as a balanced cost-quality trade-off. A detailed comparison of the pricing, inference time and rate limits is summarized in the \autoref{table_llms_pricing} of Appendix~\ref{appendix_llm_few_shot}. In addition, we observed some mysterious drops in performance when predicting spherical shape. More specifically, Mistral-medium and GPT-4-turbo produced the accuracy of 0.38 and 0.44, respectively, which dramatically decreased their average scores, while the other models under identical experimental conditions coped with the problem reasonably well.

\begin{table*}[tb]
\caption{Top performance achieved by the LLMs in prediction of nanomaterial shapes. Average accuracy corresponds to the following prompting strategy: only target classes in prompt, syntheses presented in the textual format, number of training examples $N=8$. Only accuracy is given since the corresponding data is balanced. }
\vskip 0.15in
\begin{center}
\begin{small}
\begin{tabular}{@{}lcccccc@{}}
\toprule
                   & Mistral-medium & Mistral-small & Mistral-tiny & GPT-3.5-turbo & GPT-4     & GPT-4-turbo \\ \midrule
Cube               & 0.70±0.11      & \textbf{0.76±0.08}     & 0.76±0.19    & 0.69±0.18     & 0.71±0.05 & 0.60±0.15   \\
Stick              & \textbf{0.71±0.04}      & 0.67±0.11     & 0.71±0.10    & 0.62±0.16     & 0.68±0.05 & 0.61±0.13   \\
Sphere             & 0.38±0.12      & 0.77±0.18     & 0.62±0.24    & 0.63±0.15     & \textbf{0.88±0.05} & 0.44±0.12   \\
Flat               & 0.89±0.08      & \textbf{0.92±0.07}     & 0.81±0.17    & 0.90±0.06     & 0.90±0.10 & 0.91±0.06   \\
Amorphous          & 0.70±0.15      & 0.88±0.08     & 0.53±0.16    & 0.80±0.13     & 0.87±0.12 & \textbf{0.88±0.08}   \\ \midrule
Average &   \multirow{2}{*}{0.68±0.10}        &   \multirow{2}{*}{0.80±0.10}   &      \multirow{2}{*}{0.69±0.17}     &     \multirow{2}{*}{0.73±0.13}      & \multirow{2}{*}{\textbf{0.81±0.07}} &    \multirow{2}{*}{0.69±0.11}      \\
accuracy &           &      &           &          &  &          \\ \bottomrule
\end{tabular}
\end{small}
\end{center}
\vskip -0.1in
\label{table_llms_comparison}
\end{table*}

Analyzing the impact of sampling methods and data formats (\autoref{table_prompting_strategy} shows results for one of the GPT-4 experiments), we came to the following conclusions. First, including more examples from the training set belonging to the same class as the test sample definitely benefits the prediction. We observed improvements in accuracy in all related cases. Second, textual and tabular data formats performed similarly. However, textual format consistently resulted in a 4\% increase in average accuracy, which was expected due to the nature of LLMs.

\begin{figure}[h]
	\centering
	\includegraphics[width=1\columnwidth]{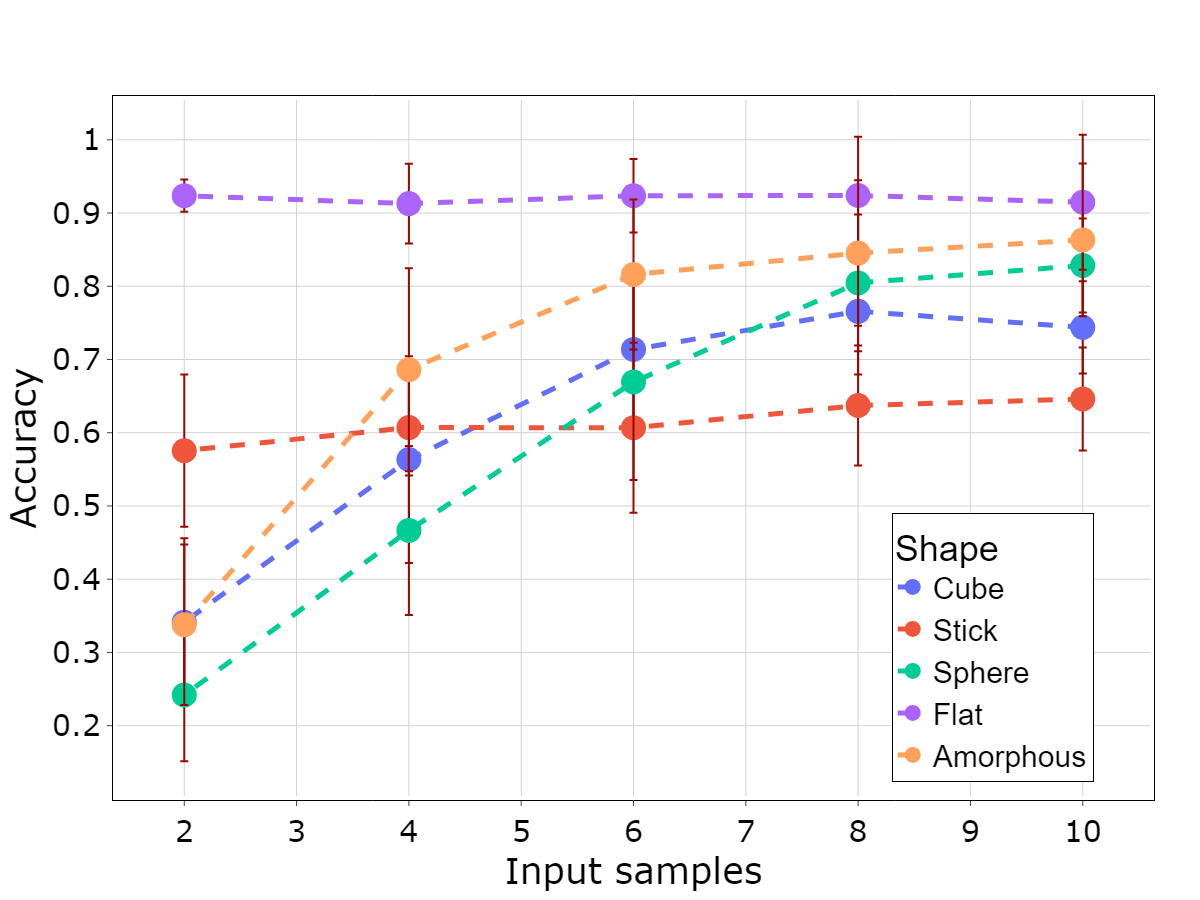} 
	\caption{Average accuracy of GPT-4 for different number of samples in prompt taken from the training set. Sampling method: only target classes in prompt. Syntheses presented in the textual format. Colors correspond to different shapes of nanoparticles.}
	\label{figure_input_samples}
\end{figure}

Finally, the number of training samples in the prompt also correlated with the performance metrics (\autoref{figure_input_samples}). For all shapes except the cube, we observed an increase in accuracy as more examples from the training set were included for prediction. However, longer prompts are also known to trigger hallucination. On top of that, there is a hard limit on the maximum prompt size for many models. Therefore, for any particular application, one has to seek another trade-off between the number of training samples and the total prompt size. In our case, the performance seemed to reach a plateau with 8 samples (see Appendix~\ref{appendix_llm_few_shot} for more details). The same configuration demonstrated the overall top performance (\autoref{table_llms_comparison}).

\begin{table}[h]
	\caption{Average accuracy for different prompting strategies of GPT-4 with $N=4$ training examples: combinations of sampling methods and data formats. Only accuracy is given since the corresponding data is balanced. }
    \vskip 0.15in
    \begin{center}
    \begin{small}
    \resizebox{0.48\textwidth}{!}{
\begin{tabular}{@{}lcccc@{}}
\toprule
Sampling               &  \multicolumn{2}{c}{At least one target}    & \multicolumn{2}{c}{Only target}                                                  \\
method               &  \multicolumn{2}{c}{class in prompt}     & \multicolumn{2}{c}{classes in prompt}                                                        \\ \midrule
Data& \multirow{2}{*}{Textual}                                    & \multirow{2}{*}{Tabular}                                 & \multirow{2}{*}{Textual}                                    & \multirow{2}{*}{Tabular}                                   \\
format &                                    &                                  &                                     &                                   \\ \midrule
Cube               & 0.52±0.10                              & 0.49±0.13                             & 0.56±0.16                               & \textbf{0.67±0.11}                              \\
Stick              & 0.54±0.03                              & 0.59±0.05                             & \textbf{0.61±0.10}                               & \textbf{0.61±0.05}                              \\
Sphere             & 0.43±0.16                              & 0.43±0.14                             & \textbf{0.54±0.08}                               & 0.32±0.10                              \\
Flat               & 0.88±0.11                              & 0.86±0.08                             & 0.92±0.05                               & \textbf{0.94±0.02}                              \\
Amorphous          & 0.57±0.20                              & 0.37±0.15                             & \textbf{0.68±0.07}                               & 0.59±0.15                              \\ \midrule
Average            & 0.59±0.12                              & 0.55±0.11                             & \textbf{0.66±0.09}                               & 0.62±0.09                             \\ \bottomrule
\end{tabular}}
\end{small}
\end{center}
\label{table_prompting_strategy}
\vskip -0.1in
\end{table}

Achieving state-of-the-art performance for nanomaterial morphology prediction with LLMs is very exciting for several reasons. First, it makes it possible for domain experts and experimentalists to avoid implementing complex data engineering pipelines and optimizing machine learning models, and use natural language to obtain the predictions instead. Second, it is obvious from our empirical results (\autoref{table_llms_comparison}) that an ensemble of LLMs would by far outperform the best classical ensemble models. Third, based on our empirical results, LLMs look especially advantageous in classification of underrepresented classes, or in small data scenarios. In particular, GPT-4 demonstrated a significant increase in accuracy when predicting less represented spherical, flat and amorphous nanoparticles (\autoref{table_rf_shapes}). Altogether, our results look very promising for the broader adoption of LLMs in the nanomaterial science.


\subsection{Text-to-image system}
Prediction of a nanoparticle shape as a categorical variable based on the selected set of properties describing the synthesis procedure is inherently subject to information loss. Intuitively, images are much better representations of shapes than any handcrafted categories, and the full text of a nanoparticle synthesis carries more information compared to a set of numerical features extracted from it. Therefore, a text-to-image paradigm previously explored in general-purpose applications \cite{rombach_high-resolution_2021} and other domains \cite{khwaja_cell-e_2022} looks appealing in the context of our problem. In the following, we attempt to prototype such a system to explore its potential despite the hard constraints on the sample size.

\begin{figure}[h]
	\centering
	\includegraphics[width=1\columnwidth]{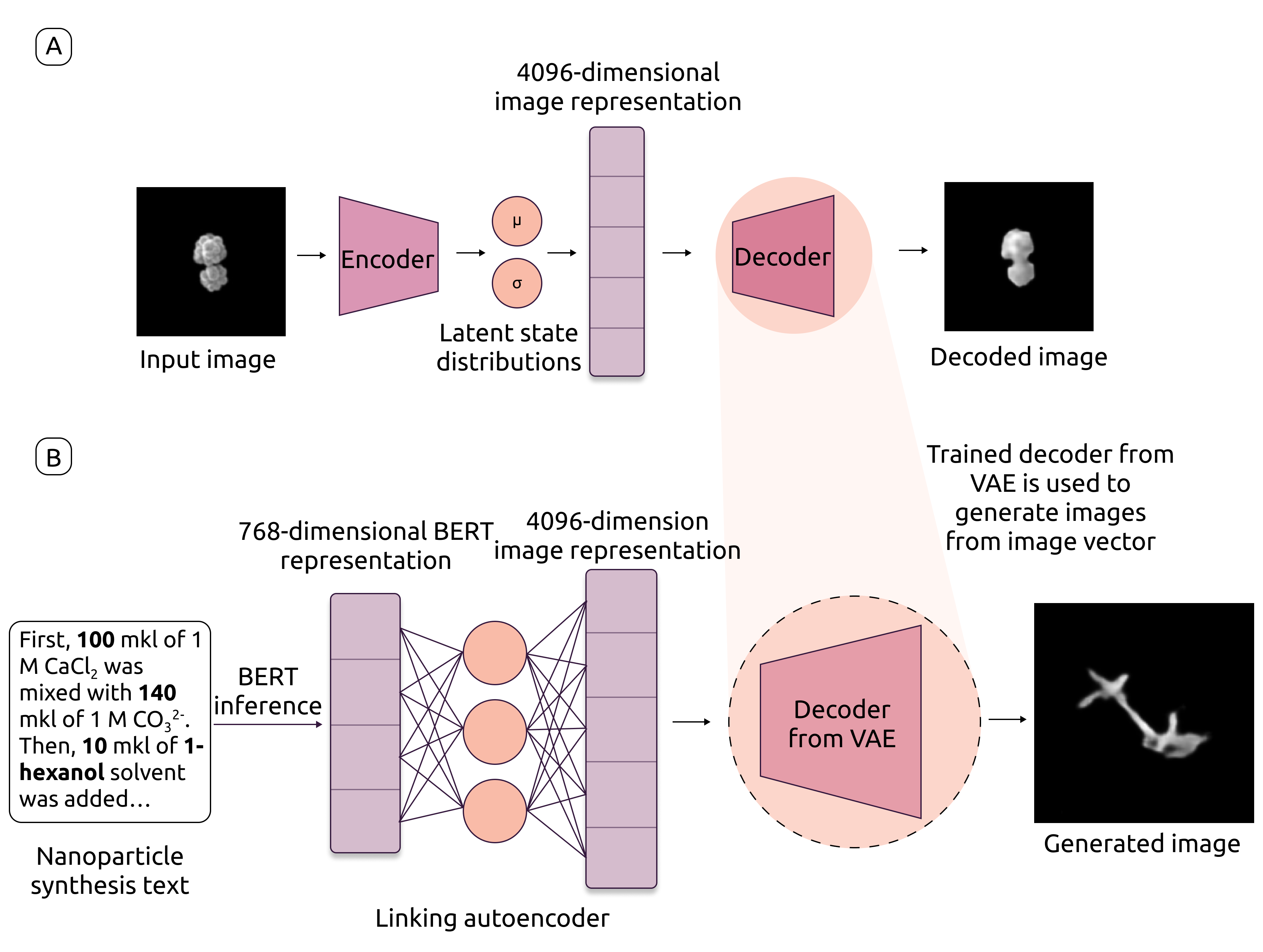} 
	\caption{A schematic of the text-to-image system prototype. A) VAE training. The images of nanoparticles are used to train a variational autoencoder (VAE). B) Final model inference. The corresponding synthesis procedures are converted into vector representations with a pretrained BERT (bottom left). The “linking” autoencoder is trained to map text and image representations (bottom center). Finally, the decoder of the VAE is used to generate new images of nanomaterials based on the descriptions of syntheses (bottom right).}
	\label{figure_generative}
\end{figure}

We break down the text-to-image system into three main components. The first one is the natural language processing model converting the text of a synthesis procedure to a vector of numerical features. The second component is the generative model with an encoder-decoder architecture designed to learn representations of images of nanoparticles. Finally, the third component is the “linking” model translating the text representations into the image representations. When combined, the three components make a generative system capable of drawing the morphology of a nanomaterial based on the description of its synthesis (\autoref{figure_generative}). 

\subsubsection{Natural language processing model}


The main requirement for the NLP model used for feature extraction was the ability to retain information about the qualitative and the quantitative features of a synthesis.
In order to select the NLP model, we formulated several classification and regression tasks related to the key features of a synthesis procedure. We used the linear evaluation setup with standard metrics \cite{kolesnikov_revisiting_2019} to compare several pretrained transformer-based models. We found that the classic BERT model \cite{devlin_bert_2018} achieved perfect scores in most tasks and, therefore, used BERT as the feature extractor in the text-to-image setup (\autoref{figure_generative}). It also met the requirement of being relatively lightweight, easy to start up and use.

\subsubsection{Autoencoder-based generative model}
The most widely spread deep learning model architectures capable of generating images are generative adversarial networks (GANs) \cite{goodfellow_generative_2014}, variational autoencoders (VAEs) \cite{kingma_auto-encoding_2013}, and diffusion models \cite{rombach_high-resolution_2021, ramesh_hierarchical_2022}. We opted for a variational autoencoder as a more stable and a more suitable solution for small datasets, given the limited amount of data available for training.

The central idea of autoencoders is to learn a compressed representation of the input data while solving a data reconstruction problem. Variational autoencoders also imply a certain probabilistic distribution in the input data, which allows it to generate meaningful outputs by sampling the latent representation after the training is complete \cite{kingma_auto-encoding_2013}. In order to plug the VAE into the text-to-image system, we first trained it on the set of SEM images and then froze the decoder part (\autoref{figure_generative}). Refer to Appendix~\ref{appendix_vae_details} and~\ref{appendix_vae_metrics} for tested VAE architectures and evaluation metrics used.

\begin{figure}[h]
	\centering
	\includegraphics[width=1\columnwidth]{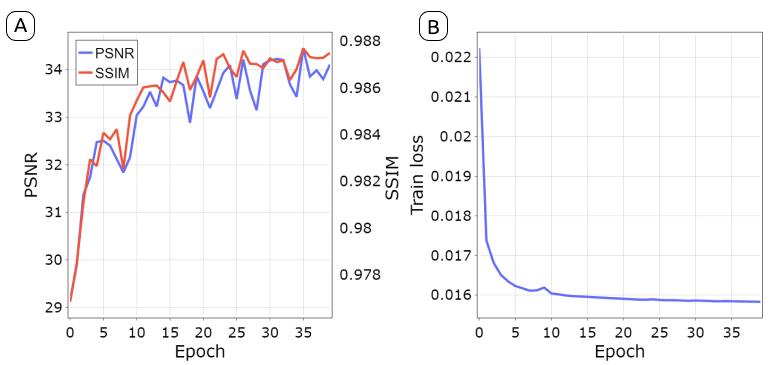} 
	\caption{A) PSNR and SSIM metrics of the selected VAE by epoch. B) Training loss of the selected VAE by epoch.}
	\label{figure_generative_metrics}
\end{figure}

We validated the final VAE model by monitoring training losses and evaluation metrics (\autoref{figure_generative_metrics}), analyzing individual examples of reconstructed images and visualizing the space of learned representations allowing to distinguish different clusters of nanoparticle shapes (Appendix~\ref{appendix_umap}).

\subsubsection{“Linking” autoencoder model}
The last component of the proposed text-to-image system is the “linking” neural network learning to map representations of the two modalities. Considering the data limitations and the empirical results described earlier, we refrained from using complex model architectures for this task. Instead, we developed another set of shallow autoencoder networks having from 3 to 8 linear layers. Like in the case of VAE, we optimized hyperparameters for each network, including the dimensionality of the latent space, to achieve the lowest reconstruction MSE (see Appendix~\ref{appendix_linking_vae} for details on training and generation phases). The best architecture for the “linking” autoencoder is given in Appendix~\ref{appendix_linking_best}.

\subsubsection{Results}


We observed that our prototype of the text-to-image system copes best with the generation of cubic nanoparticles, which was expected since the cubic shape was the most represented in the training data. For syntheses of this type of nanomaterials, the generated images were often distinct and well-shaped. It was also easier to grasp the size of cubic nanoparticles compared to other types. In general, however, the size of the dataset was insufficient to generate high quality images directly from text. Several examples of generated images are shown on \autoref{figure_generative_results} of the Appendix~\ref{appendix_linking_best}.

Despite the limited applicability of this prototype, we realized that repeated image generation based on the same synthesis parameters can provide insights into the polydispersity of NPs. Polydispersity is normally defined as $PdI=(\frac{\sigma }{2a})^{2}$, where $\sigma$ is the standard deviation of the particle diameter, and a is the mean hydrodynamic radius. We performed 50 generations of amorphous NPs with the same synthesis parameters and observed maximal diameters ranging from 30 to 80 pixels. As polydispersity characterization is critical for many applications \cite{clayton_physical_2016}, a generative model, such as the proposed prototype, could be instrumental in fast \textit{in silico} screening of NPs by estimating $PdI$ based on the predicted images.

\section{Discussion and conclusion}

In this work, we explored the potential of AI in predicting morphological properties of nanomaterials using the newly generated multimodal dataset of calcium carbonate nanoparticles. First, we investigated statistical associations between synthesis procedures and the resulting morphologies. Then, we trained and optimized tree-based ensemble models to predict multiple categories of nanomaterial shapes and sizes. After that, we systematically evaluated capabilities of the state-of-the-art LLMs in the same prediction tasks. Finally, we prototyped a text-to-image system to predict images of nanoparticles directly from the descriptions of syntheses. 

Notably, this work stands out by creating a new dataset of multiple types of nanoparticle shapes, which can be used for benchmarking in the future. This dataset opens up the possibility of predicting the shape of nanomaterials as it represents nanomaterials of multiple shapes within a single chemical system. Also, to our knowledge, we are the first to train machine learning models to distinguish between nanomaterial shapes based on the synthesis parameters. Despite these achievements, there are still several unresolved issues in the field and certain limitations to the proposed models. A more detailed discussion and a comparison with previous works are offered in Appendix~\ref{appendix_discussion} and Appendix~\ref{appendix_comparison}, respectively. 

While text-to-image applications remain largely infeasible due to the limited data availability, we identified a huge potential for future LLM applications. Not only did we observe on par performance with the classical ensemble models, we also managed to collect evidence for the superior performance of LLMs, especially in the small data scenarios. Ensemble methods for LLMs now look as a promising research direction, as less computationally expensive models like Mistral-small approach the market leader's performance in the domain-specific tasks.

\section{Data and code availability}

All datasets, scripts and results described in this work are available for reproducibility and possible transfer learning applications in this repository: \url{https://github.com/acid-design-lab/Nanomaterial_Morphology_Prediction}.

\section*{Acknowledgements}

The authors would like to thank the reviewers for their valuable feedback. The work was financially supported by the Priority 2030 Federal Academic Leadership Program.

\section*{Impact statement}

Modern machine learning techniques are evolving rapidly and sometimes find their applications almost immediately. However, in the experimental fields of science, such as materials science, the introduction of new computational methods is rather slow. Therefore, the primary goal of our work was to showcase possible applications of the most recent advances in machine learning to the design of nanomaterials, and bridge the gap between the experimentalists and the machine learning experts. There are many potential societal consequences of our work, none which we feel must be specifically highlighted here.

\bibliography{example_paper}
\bibliographystyle{icml2024}

\newpage
\appendix
\onecolumn
\section{Appendix}
\subsection{Details on statistical testing}
\label{appendix_statistics}
The Kolmogorov-Smirnov test \cite{smirnov_estimate_1939}. If the null hypothesis is rejected, this test indicates that the two samples are not drawn from the same distribution, that is, the two samples of a synthesis parameter differ in the presence or absence of a particular class of nanoparticles.
Let \((X_1^1, X_2^1, …, X_m^1)\) be independent, identically distributed real values of a parameter of a synthesis that produces cubic nanoparticles with the common cumulative distribution function \(F_{1,n}\). Let \((X_1^2, X_2^2, …, X_m^2)\) be independent, identically distributed real values of the same parameter of a synthesis which always results in nanoparticles of different shapes with the common cumulative distribution function \(F_{2,m}\). The Kolmogorov–Smirnov statistic in this case is: \(D_{n,m}  = sup_x |F_{1,n}(x^1) - F_{2,m} (x^2)|\), where \(sup\) is the supremum function. 

The null hypothesis is that the two samples are from the same continuous distribution. The null hypothesis is rejected at level \(\alpha = 0.05\) if \(D_{n,m}>\sqrt{-ln(\alpha/2)\cdot(1 + m/n)/2m}\). We applied this test for each of the real-valued parameters of synthesis and each type of nanomaterial shape and used the Bonferroni correction method similarly to the previous tests.
The results of this test were similar to those of the previous two, except that in the case of stick-shaped nanoparticles, the dependence was observed on the parameter characterizing the mass of the polymer rather than its concentration, which is not surprising given the similar nature of these two parameters.

ANOVA \cite{marsal_introduction_1987} was used to compare distributions of continuous parameters corresponding to different shapes of nanomaterials. Let \((X_1^i, X_2^i, …, X_n^i)\) be independent, identically distributed real values of a parameter of a synthesis that produces nanoparticles of a specific shape with the common cumulative distribution function \(F_{i,n}\)with the mean \(\overline{X}^i\). The formula for the one-way ANOVA F-test statistic is: \(F = \frac{\sum_{i=1}^{K}n_i(\overline{X}^i - \overline{X})^2 / (K-1)}{\sum_{i=1}^{K}\sum_{j=1}^{n_i}(X_j^i - \overline{X}^i)^2 / (N-K)}\), where \(\overline{X}^i\) denotes the sample mean in the i-th group, \(n_i\) is the number of observations in the i-th group, \(\overline{X} \) denotes the overall mean of the population, and \(K\) denotes the number of groups, where \(X_j^i\) is the \(j^{th}\) observation in the \(i^{th}\) out of \(K\) groups and \(N\) is the overall sample size. 
The null hypothesis can be formulated as follows: \(\overline{X}^i= \overline{X}^j\), for each two groups \(i\) and \(j\). If F-statistic is greater than critical p-value (at the significance level \(\alpha = 0.05\)), then the null hypothesis is rejected and distributions of this synthesis parameter in the case of at least two different shapes are different. We applied this test for each of the real-valued parameters of synthesis and each type of nanomaterial shape and used the Bonferroni correction method similarly to the previous tests.
The results of this test were consistent with the results of the first two tests, except that it failed to confirm the relationship between the shape of nanoparticle and polymer mass, although polymer concentration was still a significant parameter. All major associations between features of the synthesis and the corresponding shapes of nanoparticles are presented in \autoref{table_statistics}.

\begin{table}[h]
\caption{Significant associations between features of the synthesis and the corresponding shapes of nanoparticles. The table shows the parameters that turned out to be determinant in the synthesis of nanomaterials of one or another shape. For continuous features, the following tests were used: Mann–Whitney U test, Kruskal–Wallis H test, Kolmogorov–Smirnov test, ANOVA. Fisher exact and Chi-squared tests was used for categorical features.}
\vskip 0.15in
\begin{center}
\begin{small}
\begin{tabular}{@{}clclll@{}}
\toprule
\multicolumn{1}{l}{Shape}                  & Stick-shaped          & \multicolumn{1}{l}{Spherical}                        & Flat                                   & Cubic                 & Amorphous                              \\ \midrule
\multirow{4}{*}{Continuous}  & Temperature, C        & \multirow{4}{*}{Solvent, \% vol.}                    & \multicolumn{1}{c}{\multirow{4}{*}{-}} & Polymer, \% wt.       & \multicolumn{1}{c}{\multirow{4}{*}{-}} \\
\multirow{4}{*}{features}  & Synthesis   time      &                                                      & \multicolumn{1}{c}{}                   & Temperature, C        & \multicolumn{1}{c}{}                   \\
                                      & Polymer,   \% wt.     &                                                      & \multicolumn{1}{c}{}                   & Polymer Mwt, kDa      & \multicolumn{1}{c}{}                   \\
                                      & Polymer   Mwt, kDa    &                                                      & \multicolumn{1}{c}{}                   &                       & \multicolumn{1}{c}{}                   \\
\multirow{4}{*}{Categorical}           & & \multirow{4}{*}{Myristyltrimethylammonium} & & & \\
\multirow{4}{*}{features} & Sodium dodecylsulfate & \multirow{4}{*}{bromide} & PAA                                    & Sodium dodecylsulfate & Sodium dodecylsulfate                  \\
 & PEG                   &      & PSS                                    & PAA                   & Isopropyl alcohol                      \\
                                      & PEI                   &                                                      & PVP                                    & PSS                   & tert-Butanol                           \\
                                      &                       &                                                      &                                        & Polymer absence       & Propylene glycol                      \\ \bottomrule
\end{tabular}
\end{small}
\end{center}
\vskip -0.1in
\label{table_statistics}
\end{table}

\subsection{Tree-based ensemble models}
\label{appendix_ensemble_models}
In order to achieve the best results for each of the Random Forest and Gradient Boosted Trees models, we optimized the hyperparameters for each of them, and built the models with different splits of the original dataset. The final metrics for each model were calculated by predicting at 5 different random states, after which the mean value as well as the standard deviation were calculated. Most of the functions used to prepare the dataset and use the models were implemented using the scikit-learn library \cite{scikit-learn}.

In case of the Random Forest, optimization of the following parameters was performed: \texttt{n\_estimators}, \texttt{max\_features}, \texttt{max\_depth}, \texttt{min\_samples\_leaf}, \texttt{max\_leaf\_nodes}. In case of Gradient Boosted Trees, the optimized parameters were: \texttt{gamma}, \texttt{colsample\_bytree}, \texttt{max\_depth}, \texttt{n\_estimators}, \texttt{learning\_rate}. Hyperparameter optimization was performed using 5-fold cross-validated grid-search. Given that some target classes were underrepresented, we prepared three test sets in advance for a more thorough assessment of performance. The test sets contained 33\%, 20\% or 15\% of the total number of samples.
A summary \autoref{table_appendix_ensemble_test_subset_comparison} provides motivation for testing several data splits. In our case, the lowest mean standard deviation was observed in 33\% test split for both, accuracy and F1 score, among all the experiments.
Also, for each model, the optimal threshold was found to solve the problem of class imbalance. This was achieved by balancing precision and recall metrics.

\begin{table}[h]
\caption{Comparison of different data splits. A representative test set was obtained with 33\% of total number of samples.}
\vskip 0.15in
\begin{center}
\begin{small}
\begin{tabular}{@{}ccccc@{}}
\toprule
Validation subset   & Average  & Average standard      & Average  & Average standard      \\
size of dataset, \% & accuracy & deviation of accuracy& F1 score & deviation of F1 score \\ \midrule
33                  & \textbf{0.74}     & \textbf{0.11}                  & \textbf{0.53}     & \textbf{0.09}                  \\
20                  & 0.69     & 0.13                  & 0.51     & 0.11                  \\
15                  & 0.65     & 0.18                  & 0.49     & 0.15                \\
\bottomrule 
\end{tabular}
\end{small}
\end{center}
\label{table_appendix_ensemble_test_subset_comparison}
\vskip -0.1in
\end{table}

For the best models, we also performed feature importance analysis by constructing SHAP diagrams showing the most important features in model performance. \autoref{figure_shap} below shows the 10 most important features for the Random Forest model with optimal parameters predicting the stick shaped nanomaterials. Among these features, statistical relationship with the given shape of nanomaterials was confirmed for the following features: 'Temperature, C', 'Synthesis time', 'Polymer, \% wt.', 'Polymer Mwt, kDa', 'Sodium dodecylsulfate', 'PEI'. This is an additional validation of our models, as the results of the analysis of feature importance almost completely correspond to the previously discovered statistical patterns that were presented in \autoref{table_statistics}.

\begin{figure}[h]
	\centering
	\includegraphics[width=0.5\columnwidth]{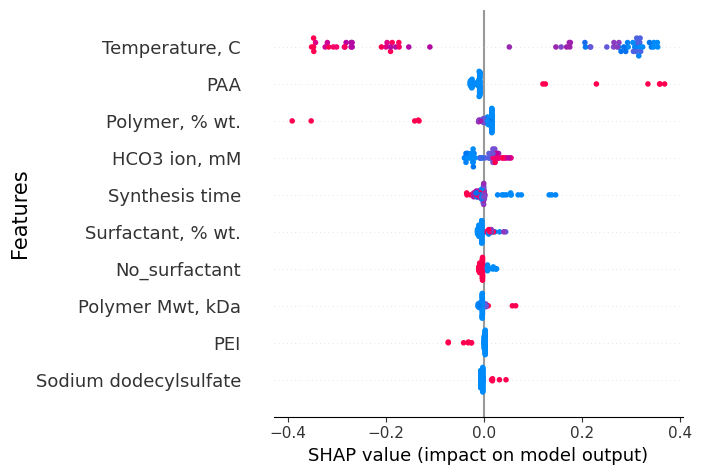} 
	\caption{Results of feature importance analysis in the form of SHAP values for the top 10 features for the best Random Forest model for predicting stick shaped nanoparticles.}
	\label{figure_shap}
\end{figure}

\subsection{Texts of synthesis procedures and prompts}
\label{appendix_synthesis_texts}
In order to make predictions of the morphology of nanomaterials based on their synthesis text using LLMs, special templates were created, which were then used to be filled with parameters for a particular synthesis. From these, the final textual prompt for LLM was compiled. These templates were similarly used in the development of a generative text-to-image system. Two examples of such templates are given below.

Template example 1:

\textit{“Synthesis was carried out using the co-precipitation technique. Initially, {ca\textunderscore conc} mkl of 1 M CaCl2 was combined with {pol\textunderscore vol} mkl of {pol\textunderscore conc} \% wt. {polymer} polymer having a molecular weight of {pol\textunderscore mass} kDa. Subsequently, {solvent\textunderscore volume} mkl of {solvent} was introduced, and the volume adjusted to 500 mkl using distilled water. Following that, {co3\textunderscore conc} mkl of 0.1 M Na2CO3 was mixed with {hco3\textunderscore conc} mkl of 0.1 M NaHCO3, along with {surf\textunderscore vol} mkl of {surf\textunderscore conc} \% wt. {surfactant} serving as the surfactant. Another {solvent\textunderscore volume} mkl of {solvent} was added, and the volume adjusted to 500 mkl using distilled water. Two resulting solutions, both heated to {r\textunderscore temp} C prior to the reaction, were combined under continuous stirring at {stir\textunderscore ratio} rpm while maintaining the temperature. The reaction proceeded for {r\textunderscore time} min, followed by centrifugation.”}

Template example 2:

\textit{"All materials were synthesized via the co-precipitation technique. In the first step, {ca\textunderscore conc} mkl of 1 M CaCl2 was combined with {pol\textunderscore vol} mkl of {pol\textunderscore conc} \% wt. {polymer} polymer, characterized by a molecular weight of {pol\textunderscore mass} kDa. This was followed by the addition of {solvent\textunderscore volume} mkl of {solvent}, and the volume was adjusted to 500 mkl using distilled water. In the subsequent step, {co3\textunderscore conc} mkl of 0.1 M Na2CO3, {hco3\textunderscore conc} mkl of 0.1 M NaHCO3, and {surf\textunderscore vol} mkl of {surf\textunderscore conc} \% wt. {surfactant} surfactant were combined. Once more, {solvent\textunderscore volume} mkl of {solvent} was added, and the volume was adjusted to 500 mkl using distilled water. Finally, two solutions, both heated to {r\textunderscore temp} C before the reaction, were mixed under stirring at {stir\textunderscore ratio} rpm while maintaining the temperature. The reaction proceeded for {r\textunderscore time} min, followed by centrifugation."}

Below is also one of the text-based prompts that were given to the model before predicting the morphology of nanomaterials on the test subset. 

\textit{"You are an expert in the synthesis of nanomaterials. You analyze the conditions for obtaining a nanomaterial and predict what particle shapes will be present in the synthesized material. There are five particle shapes: 'Cube', 'Stick', 'Sphere', 'Flat' and 'Amorphous'. A nanomaterial can contain particles of different shapes. If you cannot say exactly what it is, list the forms that have the highest probability in those conditions.} 

\textit{CaCO3 nanoparticles were synthesized by the co-precipitation approach according to the following manner. In separate burettes two solutions were made, 57 mkl of 1 M CaCl2 and 20 mkl of 0.155 \% wt. PEI with molecular weight of 25.0 kDa were mixed in 200.0 mkl of 1-Hexanol before dilution with distilled water up to 500 mkl. Similarly, 140 mkl of 0.1 M Na2CO3 and 200 mkl of 0.1 M of NaHCO3 were combined with 20 mkl of 0.43 \% wt. Myristyltrimethylammonium bromide and 200.0 mkl of 1-Hexanol. Then, the solution was also diluted in 500 mkl of water. Both solutions were heated up to 68 C right before mixing under stirring at 1000 rpm for 8 min 0 sec min following centrifugation.}

\textit{Answer: 'Cube, Stick'"}

An example is also given for the case of prompts that used tabular data. In this case, only the way the synthesis was presented differed, but the overall structure of the prompt remained the same.

\textit{"You are an expert in the synthesis of nanomaterials. You analyze the conditions for obtaining a nanomaterial and predict what particle shapes will be present in the synthesized material. There are five particle shapes: 'Cube', 'Stick', 'Sphere', 'Flat' and 'Amorphous'. A nanomaterial can contain particles of different shapes. If you cannot say exactly what it is, list the forms that have the highest probability in those conditions.} 

\textit{Ca ion, mM: 148; CO3 ion, mM: 0; HCO3 ion, mM: 100; Polymer Mwt, kDa: 0.0; Polymer, \% wt.: 0.0; Surfactant, \% wt.: 0.0; Solvent, \% vol.: 0.0; Stirring, rpm: 0; Temperature, C: 31; Synthesis time: 129; Hexadecyltrimethylammonium bromide: 0; Myristyltrimethylammonium bromide: 0; No\_surfactant: 1; Sodium dodecylsulfate: 0; Triton X-100: 0; 1-Hexanol: 0; Dimethylformamide: 0; Ethylene glycol: 0; Isopropyl alcohol: 0; Methyl alcohol: 0; No\_solvent: 1; Propylene glycol: 0; tert-Butanol: 0; No\_polymer: 1; PAA: 0; PEG: 0; PEI: 0; PSS: 0; PVP: 0}

\textit{Answer: 'Flat'"}

\subsection{Few-shot classification}
\label{appendix_llm_few_shot}

To optimize the number of input examples and the proportion of the test subset, experiments were conducted with the GPT-4 model for the text subset. The table below summarizes these results (\autoref{table_input_samples}). The low impact of the proportion of the test subset is obvious, but the number of input examples has a significant impact on the metrics.

\begin{table}[h]
	\caption{Average accuracy of GPT-4 for different number of input samples in prompt $N$ taken from the training set. Sampling method: only target classes in prompt. Syntheses presented in textual format.}
    \vskip 0.15in
    \begin{center}
    \begin{small}
\begin{tabular}{@{}ccccccccc@{}}
\toprule
\multirow{2}{*}{Input}           & \multicolumn{1}{c}{\multirow{2}{*}{Test}} & \multicolumn{5}{c}{Shape}              &          \multirow{2}{*}{Average}                      \\ \cline{3-7}
\multirow{2}{*}{samples}         & \multicolumn{1}{c}{\multirow{2}{*}{subset size}}                            & \multirow{2}{*}{Cube}      & \multirow{2}{*}{Stick}     & \multirow{2}{*}{Sphere}    & \multirow{2}{*}{Flat}      & \multirow{2}{*}{Amorphous} &  \multirow{2}{*}{accuracy}  \\ 
  &    &   &   &   &   &   &   \\ \midrule
\multirow{3}{*}{2}               & 0.15                                            & 0.29±0.14 & 0.58±0.15 & 0.27±0.10 & 0.92±0.02 & 0.33±0.12 & 0.48±0.10 \\
                                 & 0.2                                             & 0.40±0.09 & 0.56±0.12 & 0.20±0.10 & 0.92±0.03 & 0.38±0.15 & 0.49±0.10 \\
                                 & 0.33                                            & 0.33±0.12 & 0.59±0.05 & 0.25±0.07 & 0.93±0.02 & 0.30±0.07 & 0.48±0.07 \\ \midrule
\multirow{3}{*}{4}               & 0.15                                            & 0.54±0.15 & 0.64±0.05 & 0.38±0.14 & 0.93±0.01 & 0.64±0.23 & 0.63±0.12 \\
                                 & 0.2                                             & 0.59±0.13 & 0.56±0.06 & 0.49±0.12 & 0.89±0.09 & 0.73±0.12 & 0.65±0.10 \\
                                 & 0.33                                            & 0.56±0.15 & 0.62±0.09 & 0.54±0.09 & 0.92±0.05 & 0.68±0.07 & 0.66±0.09 \\ \midrule
\multirow{3}{*}{6}               & 0.15                                            & 0.76±0.08 & 0.65±0.11 & 0.63±0.14 & 0.94±0.00 & 0.79±0.14 & 0.75±0.09 \\
                                 & 0.2                                             & 0.67±0.08 & 0.56±0.13 & 0.68±0.17 & 0.90±0.08 & \textbf{0.88±0.03} & 0.74±0.10 \\
                                 & 0.33                                            & 0.71±0.17 & 0.61±0.11 & 0.70±0.10 & 0.93±0.07 & 0.78±0.14 & 0.74±0.12 \\ \midrule
\multirow{3}{*}{8}               & 0.15                                            & \textbf{0.84±0.07} & 0.62±0.11 & 0.81±0.10 & \textbf{0.96±0.03} & 0.84±0.08 & \textbf{0.81±0.08} \\
                                 & 0.2                                             & 0.72±0.08 & 0.67±0.05 & 0.80±0.10 & 0.91±0.08 & 0.87±0.10 & 0.80±0.08 \\
                                 & 0.33                                            & 0.74±0.11 & 0.63±0.08 & 0.80±0.08 & 0.90±0.13 & 0.83±0.12 & 0.78±0.10 \\ \midrule
\multirow{3}{*}{10}              & 0.15                                            & 0.78±0.06 & 0.59±0.11 & 0.84±0.07 & 0.94±0.03 & \textbf{0.88±0.07} & \textbf{0.81±0.07} \\
                                 & 0.2                                             & 0.71±0.05 & \textbf{0.68±0.05} & \textbf{0.88±0.05} & 0.90±0.10 & 0.87±0.12 & \textbf{0.81±0.07} \\
                                 & 0.33                                            & 0.74±0.07 & 0.66±0.06 & 0.77±0.07 & 0.90±0.15 & 0.84±0.12 & 0.78±0.09 \\ \bottomrule
\end{tabular}
\end{small}
\end{center}
\vskip -0.1in
\label{table_input_samples}
\end{table}

\begin{table}[h]
\caption{Comparison of computational resources of LLMs: time per one complete experiment (in case of text dataset, an average prompt was around 3000 tokens), price per 1M tokens in USD, limits for requests per minute and 1000 tokens per minute.}
\vskip 0.15in
\begin{center}
\begin{small}

\begin{tabular}{@{}lcccccc@{}}
\toprule
                                                                          & Mistral-medium & Mistral-small & Mistral-tiny & GPT-3.5-turbo & GPT-4 & GPT-4-turbo \\ \midrule
\begin{tabular}[c]{@{}l@{}}Time per \\ experiment, s\end{tabular}         & 46             & 21            & 19           & 44            & 63    & 53          \\ \midrule
\begin{tabular}[c]{@{}l@{}}Input price per \\ 1M tokens, USD\end{tabular} & 2.7            & 0.7           & 0.1526       & 0.5           & 30    & 10          \\ \midrule
\begin{tabular}[c]{@{}l@{}}Tokens \\ limit, 1000/min\end{tabular}         & 2000           & 2000          & 2000         & 60            & 10    & 150         \\ \midrule
\begin{tabular}[c]{@{}l@{}}Requests \\ limit, 1/min\end{tabular}          & 120            & 120           & 120          & 500           & 500   & 500       \\ \bottomrule 
\end{tabular}
 
\end{small}
\end{center}
\vskip -0.1in
\label{table_llms_pricing}
\end{table}

\subsection{VAE: implementation details}
\label{appendix_vae_details}

We have experimented with several ResNet architectures but also developed a few custom architectures for the VAE. ResNet is the classical convolutional neural network originally proposed for the classification tasks \cite{he_deep_2015}. It consists of several blocks of convolutional, batch normalization and ReLU layers, and several depth options are available. Jens Behrmann et al. showed that invertible ResNets can also be used as generative models \cite{behrmann_invertible_2018} and, therefore, we used the reversed ResNet from PyTorch Lightning Bolts\footnote[1]{https://lightning-bolts.readthedocs.io/en/latest/} as a decoder for the VAE. Additionally, we developed several shallow networks varying the number of convolutional blocks and the dimensionality of the bottleneck layer as custom VAE architectures.

We trained all the architectures in the grid search setup optimizing several hyperparameters, such as batch size, learning rate, Kullback-Leibler (KL) divergence coefficient and image size, to achieve the lowest BCE loss. 

Based on the training losses and the evaluation metrics described in Appendix~\ref{appendix_vae_metrics}, we selected one of the custom architectures as the best. We failed to achieve on-par performance with the ResNet backbones, likely due to insufficient number of training examples. The top-performant VAE architecture had only 4 convolutional blocks for the encoder and 4 upsampling blocks for the decoder with 4096 dimensions in the latent space. The optimal set of hyperparameters was 128×128 for the image size, 64 for the batch size, 0.001 for the learning rate, and 0.01 for the KL divergence coefficient. The corresponding training curves are depicted on \autoref{figure_generative_metrics}.

\subsection{VAE: metrics}
\label{appendix_vae_metrics}

For each combination of hyperparameters, the trained VAEs were evaluated on the test set. Two metrics reflecting the similarity of the original and the decoded images were used to compare architectures: structural similarity index measure (SSIM) \cite{wang_image_2004} and peak signal-to-noise ratio (PSNR) \cite{fardo_formal_2016}. SSIM is a standardized measure of the difference between the compressed and the original image, ranging from -1 to 1. It is defined by the following formula:

\[
SSIM(X,Y)=\frac{(2\mu _{x}\mu _{y} + c_{1})(2\sigma _{xy} + c_{2})}{(\mu _{x}^{2}+\mu_{y}^{2}+c_{1})(\sigma_{x}^{2}+\sigma_{y}^{2}+c_{2})}
\]
where $X$, $Y$ are the images, $\mu_x, \mu_y$ are the mean pixel values of the images, $\sigma_x, \sigma_y$ are the variances of the pixel values, $\sigma_{xy}$ is the covariance, and $c_{1}, c_{2}$ are the coefficients stabilizing the division. PSNR is a simpler metric showing the ratio of the contribution of the maximum value of the original image to the contribution of noise in the compressed image. This metric is calculated using the following formula:

\[
PSNR(X,Y)=20log_{10}(\frac{max(X)}{\sqrt{e(X,Y)}})
\]
where $X$ is the original image of size m×n, $Y$ is the compressed image of the same size, and $e(X,Y)=\frac{1}{mn}\sum_{j=1}^{m}\sum_{i=1}^{n}(x_{ji}-y_{ji})^{2}$ is the mean squared deviation between the pixels of two images.

\subsection{“Linking” VAE: training and generation phases}
\label{appendix_linking_vae}

The training process was organized into the following key steps. For each training example:
\begin{enumerate}
	\item Choose a text template of a synthesis procedure randomly and fill in the corresponding experimental parameters.
	\item Obtain text representations with a pretrained BERT.
	\item Obtain image representations with a pretrained VAE.
	\item Perform a forward pass to convert text representations into image representations.
	\item Calculate the loss and backpropagate the error.
\end{enumerate}

After the training is done, morphology of a new nanomaterial described in a synthesis procedure can be predicted as follows:
\begin{enumerate}
	\item Obtain representations of the synthesis procedure text with a pretrained BERT.
	\item Apply the “linking” autoencoder to predict the corresponding image representations.
	\item Apply the decoder part of the VAE to predict the image of the nanomaterial.
\end{enumerate}

\subsection{“Linking” VAE: best architecture}
\label{appendix_linking_best}

\begin{figure}[h]
	\centering
	\includegraphics[width=0.8\columnwidth]{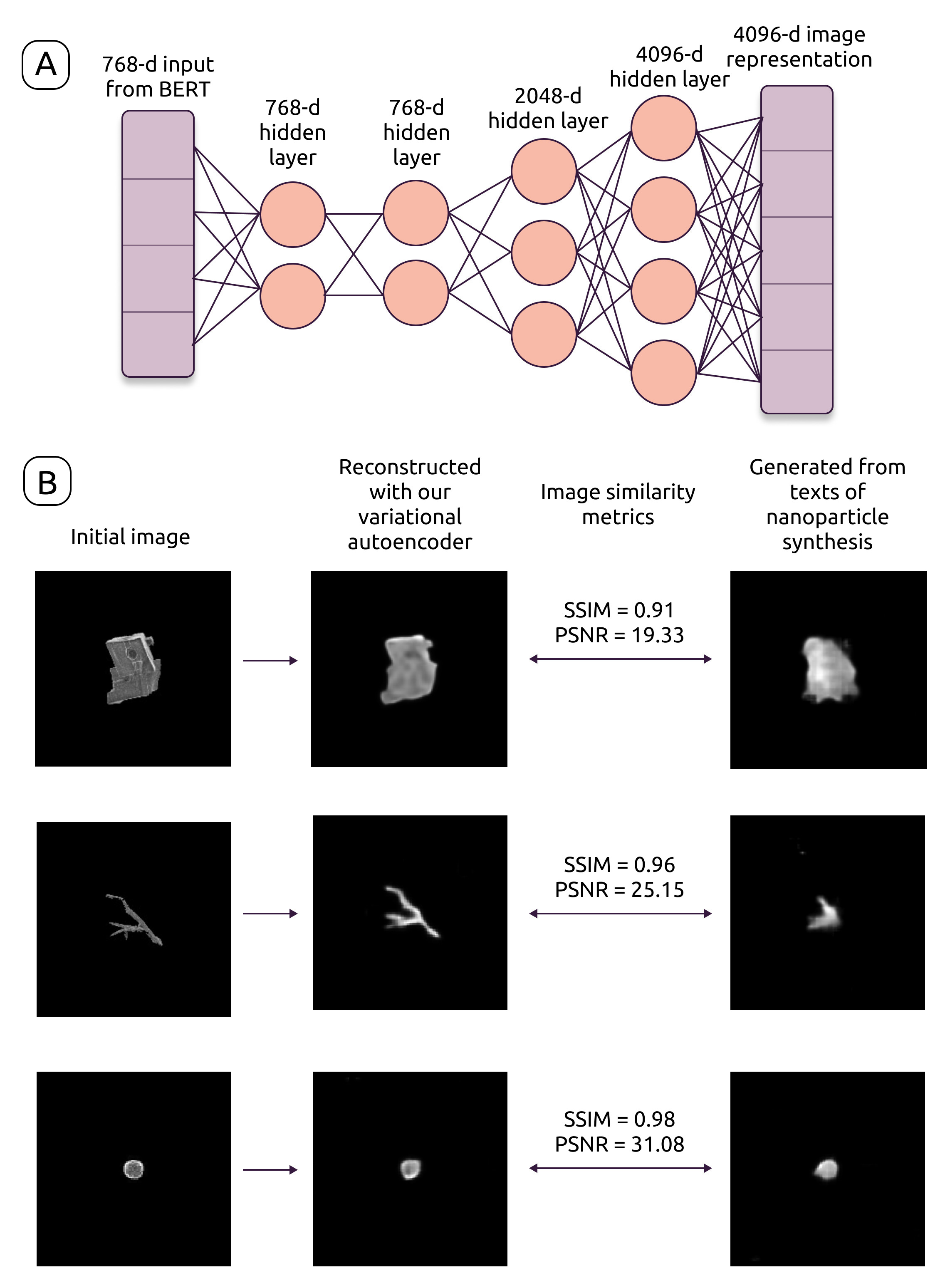} 
	\caption{A) The encoder-decoder architecture of the linking neural network. B) Comparison of real images to their VAE reconstructions and the images generated from the corresponding synthesis texts.}
	\label{figure_generative_results}
\end{figure}

The final architecture of the “linking” VAE is shown on \autoref{figure_generative_results}A. It consists of 4 linear layers and has 768-dimensional latent space. The optimal hyperparameters were 8 for the batch size, 0.00001 for the learning rate. \autoref{figure_generative_results}B shows individual examples of nanoparticles reconstructed and generated from texts. Three different shapes are given.

\subsection{Data visualization}
\label{appendix_umap}

\begin{figure}[h]
	\centering
	\includegraphics[width=1\columnwidth]{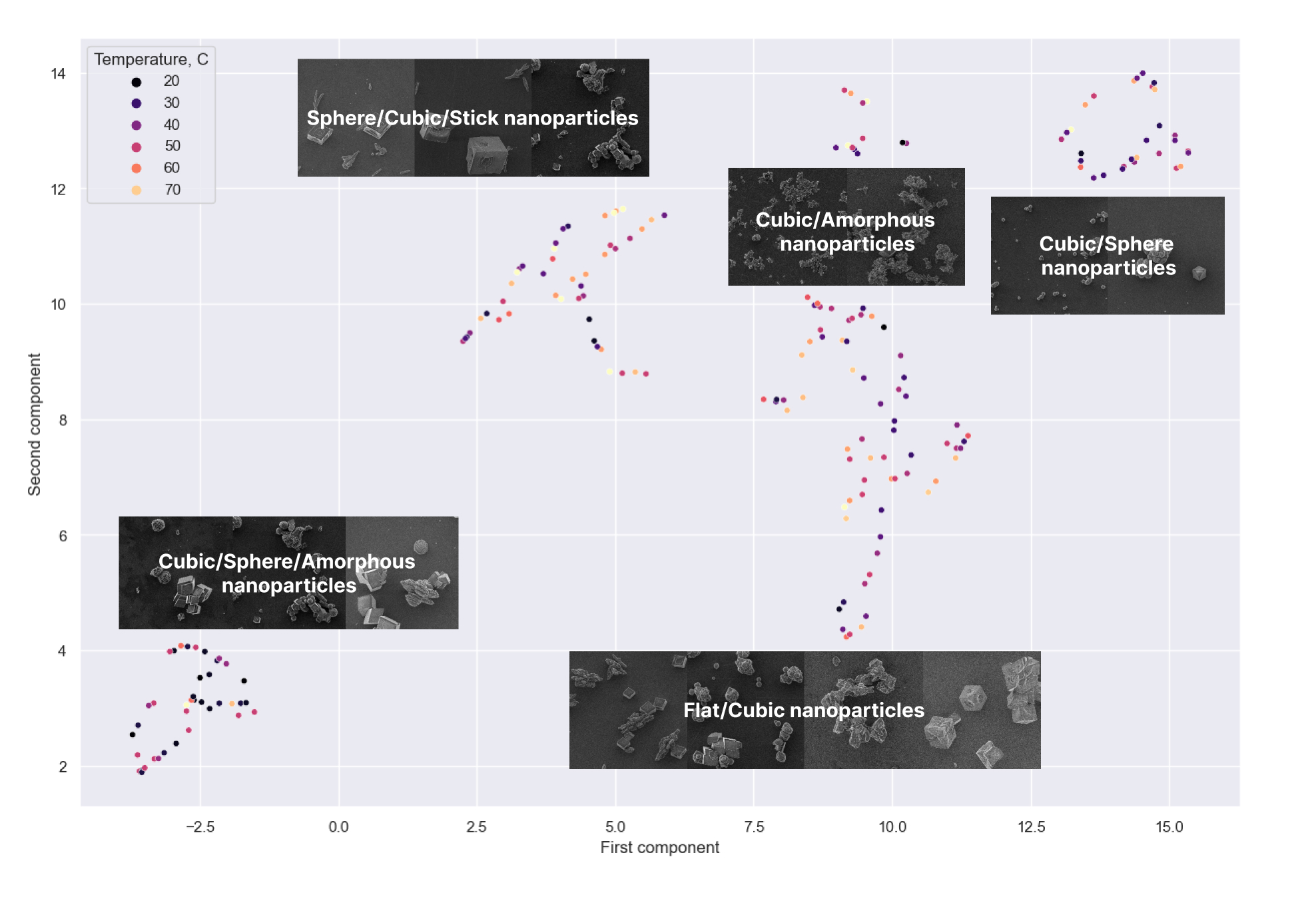} 
	\caption{Representation of the latent space of the variational autoencoder trained on our image dataset. Colors indicate the shapes of the nanomaterials, and the axes are the UMAP components after dimensionality reduction.}
	\label{figure_umap}
\end{figure}

We visualized the space of learned representations of the VAE to validate the model and gain additional insights into various dependencies between the features and the target classes. For that, we used UMAP to compress the bottleneck 4096 dimensions to 2D (\autoref{figure_umap}). Each dot represents a single representative nanoparticle from one of the 215 syntheses. We observed five clusters having 2-3 particular shapes as the most prominent. Based on the literature and the statistical evaluation, we expected to see drastic differences in temperatures for different clusters. However, we could observe a single bottom-right cluster having lower synthesis temperatures on average.

\subsection{Comparison with previous works}
\label{appendix_comparison}

A comparison with the most relevant previous works is given in \autoref{table_comparison}.

\begin{table}[h]
		\caption{Comparison with other works. *a prototype of the text-to-image system}
		\label{table1}
    \vskip 0.15in
    \begin{center}
    \begin{small}
	\centering
\begin{tabular}{@{}ccccccc@{}}
\toprule Prediction  & Data  & Number  & Best         & \multirow{2}{*}{Availability}     & Generative  & \multirow{2}{*}{Reference} \\
 task & points & of shapes & metric & & design & \\
\midrule
Size            & 103         & 1                & MAPE = 0.70\%      & Only dataset     & No                &  \cite{shafaei_predictive_2020}         \\
\midrule
Size            & 98          & 1                & MAPE = 4\%         & -                & No                &     \cite{iakovlev_artificial_2019}       \\
\midrule
Size            & 26          & 1                & MAPE = 9.10\%      & -                & No                &       \cite{pellegrino_machine_2020}    \\
\midrule
Size and shape  & 215         & 5                & Accuracy = 0.93~ ~ & Code and dataset & Yes*              &    Our work       \\ \bottomrule
\end{tabular}
    \end{small}
    \end{center}
    \label{table_comparison}
    \vskip -0.1in
\end{table}

\subsection{Discussion on limitations}
\label{appendix_discussion}

Our ML models were trained to predict 5 types of nanomaterial shapes, some of which were underrepresented (\autoref{table_rf_shapes}). This limitation can be mitigated by either adjusting the prediction threshold, or oversampling techniques. Ultimately, this issue can only be resolved by expanding the dataset for underrepresented classes. In the context of the text-to-image system, we always refer to a prototype acknowledging its limitations, such as low diversity of generated images and their quality, caused by the limited training examples available. Therefore, most of the barriers to training a more universal and accurate model for prediction of nanomaterial morphology are related to insufficient quality and number of existing datasets. There is currently no unified database with syntheses and properties of different nanoparticles that is well documented and publicly available. Therefore, applied AI researchers have to resort to small single study datasets or larger datasets of a single experimental system extensively studied in the past (\autoref{table_comparison}). Both approaches impose severe limitations on machine learning and, even more so, on deep learning applications that typically require a lot more training data. Thus, a collective effort towards assembling a curated database of nanomaterials with deep characterization of their properties is long overdue.

Additional challenges arise from the data preprocessing steps dealing with SEM. Many syntheses result in numerous overlaying NPs on a single SEM image, such that it is difficult even for a human eye to distinguish between individual NP units. Since image segmentation methods have already reached quite an advanced level, we anticipate major breakthroughs rather on the experimental and the imaging technology side.

\subsection{Computing infrastructure}
\label{appendix_computing}

\begin{table*}[h]
		\caption{Computing infrastructure used for study experiments.}
		\label{table1}
    \vskip 0.15in
    \begin{center}
    \begin{small}
	\centering
\begin{tabular}{@{}ll@{}}
\toprule 
CPU              & AMD Ryzen 7 3700X 3.60 GHz 8-Core Processor \\
GPU              & NVIDIA GeForce RTX 3090 24 GB of GPU memory \\
RAM              & 32.0 GB                                     \\
Operating system & Windows 11 Pro N                            \\
Python           & 3.9                                        
       \\ \bottomrule
\end{tabular}
    \end{small}
    \end{center}
    \label{table_computing}
    \vskip -0.1in
\end{table*}

\end{document}